\documentclass{article}

\PassOptionsToPackage{numbers, compress}{natbib}
\usepackage[preprint]{neurips_2024}

\usepackage{color}

\usepackage[utf8]{inputenc} 
\usepackage[T1]{fontenc}    
\usepackage{microtype}      
\usepackage{capt-of}
\usepackage{xcolor}
 
\usepackage{wrapfig}
\usepackage{xspace}
\usepackage{amsmath,amssymb,amsfonts,dsfont,pifont,bm,bbm,mathrsfs,mathtools,nicefrac}
\usepackage{algorithm,algpseudocode,listings}
\usepackage{booktabs,multirow,adjustbox,diagbox,threeparttable}
\definecolor{citeblue}{RGB}{48,111,186}
\usepackage[pagebackref=false,breaklinks=true,colorlinks=true,citecolor=citeblue,bookmarks=false]{hyperref}
\usepackage{cleveref}  

\crefname{section}{Sec.}{Secs.}
\Crefname{section}{Section}{Sections}
\crefname{table}{Tab.}{Tabs.}
\Crefname{table}{Table}{Tables}
\crefname{figure}{Fig.}{Figs.}
\Crefname{figure}{Figure}{Figures}
\crefname{equation}{Eq.}{Eqs.}
\Crefname{equation}{Equation}{Equations}
\newcommand{\tocite}[1]{\textcolor{red}{[TO CITE]}}
\newcommand{\method}{\textcolor{red}{TODO}\xspace}

\def\ie{{\it{i.e.,~}}}
\def\eg{{\it{e.g.,~}}}

\def\method{ViViD}

\title{ViViD: \underline{Vi}deo \underline{Vi}rtual Try-on using \underline{D}iffusion Models}

\author{
  Zixun Fang$^{1,2}$\thanks{Work done when the first author interned at Alibaba Yuanjing.} \quad Wei Zhai$^{1\dag}$ \quad Aimin Su${^{2}}$ \quad Hongliang Song$^{2}$ \quad Kai Zhu$^{2}$ \\ \textbf{Mao Wang}$^{2}$ \quad \textbf{Yu Chen}$^{2}$\thanks{Wei Zhai and Yu Chen are the corresponding authors.} \quad \textbf{Zhiheng Liu}$^{1}$ \quad \textbf{Yang Cao}$^{1}$ \quad
   \textbf{Zheng-Jun Zha}$^{1}$\\
\\
  $^1$University of Science and Technology of China \quad
  $^2$Alibaba Group 
}

\begin{document}

\maketitle

\begin{abstract}

Video virtual try-on aims to transfer a clothing item onto the video of a target person. Directly applying the technique of image-based try-on to the video domain in a frame-wise manner will cause temporal-inconsistent outcomes while previous video-based try-on solutions can only generate low visual quality and blurring results. In this work, we present ViViD, a novel framework employing powerful diffusion models to tackle the task of video virtual try-on. Specifically, we design the Garment Encoder to extract fine-grained clothing semantic features, guiding the model to capture garment details and inject them into the target video through the proposed attention feature fusion mechanism. To ensure spatial-temporal consistency, we introduce a lightweight Pose Encoder to encode pose signals, enabling the model to learn the interactions between clothing and human posture and insert hierarchical Temporal Modules into the text-to-image stable diffusion model for more coherent and lifelike video synthesis. Furthermore, we collect a new dataset, which is the largest, with the most diverse types of garments and the highest resolution for the task of video virtual try-on to date. Extensive experiments demonstrate that our approach is able to yield satisfactory video try-on results. The dataset, codes, and weights will be publicly available. 
Project page: \url{https://alibaba-yuanjing-aigclab.github.io/ViViD}.

\end{abstract}

\section{Introduction}\label{sec:intro}
\begin{figure}[!h]
  \centering
    \vspace{-1.0cm}
  \includegraphics[width=0.9\linewidth]{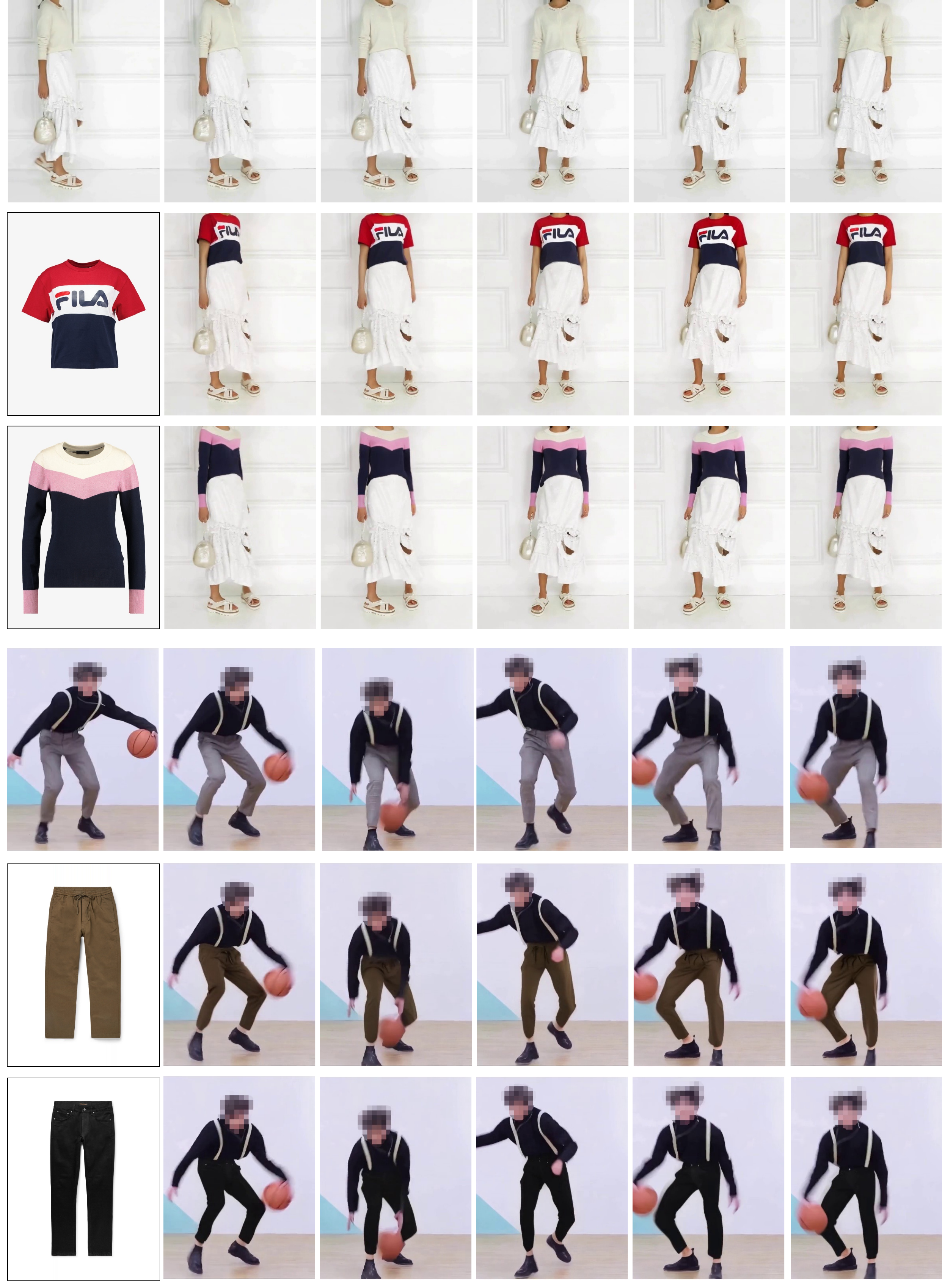}
  \caption{\small{Outfitted videos ($512 \times 384$) generated by our \method. The 1st and 4th rows are source videos. Please zoom in for more details.}}
  \label{fig:demo}
\end{figure}
With the rapid development of the internet and e-commerce industry, individuals now have facile access to various types of garments via online platforms, facilitating seamless selection and purchase processes.
Although customers can gain insight into the texture, shape, and details of clothing by viewing pictures on e-commerce websites, relying solely on these in-shop garment images makes it difficult for consumers to imagine how the clothes will actually fit and look on them, particularly in terms of how they interact with the body's movements.
The inability to physically try on the clothing before purchasing can lead to dissatisfaction and the need for returns or exchanges, adding further frustration to the online shopping experience.
Recently, the task of video virtual try-on has emerged to tackle this issue.
This task aims to synthesize a natural and realistic video that preserves the identity of the target clothes while keeping the rest of the source video unchanged.

Previous studies mainly focused on image-based virtual try-on~\cite{han2017viton,wang2018toward,baldrati2023multimodal,choi2021viton,ge2021parser,morelli2022dresscode,xu2024ootdiffusion,chen2023anydoor}.
Among them, VITON-HD~\cite{choi2021viton} introduced an ALIAS generator to preserve clothes details and successfully synthesized high-resolution images.
Dress Code~\cite{morelli2022dresscode} developed a new dataset to expand the try-on task from one single category to multiple categories.
Multimodal Garment Designer~\cite{baldrati2023multimodal} utilized latent diffusion models to receive multimodal inputs.
Despite the significant success of these methods in image virtual try-on, directly applying them to videos would result in catastrophic outcomes.
Frame-wise image try-on would lead to flickers and artifacts in the generated content due to the neglect of temporal consistency.

In recent years, some researchers have attempted to replicate the success of image try-on in the video domain~\cite{jiang2022clothformer,dong2019fw,kuppa2021shineon,zhong2021mv,xu2024tunnel}.
Despite considerable efforts, these works still suffer from issues such as flickering and spatial-temporal inconsistency.
We argue that the following reasons are key to the situation: 1). There remains a lack of a high-quality dataset.
The commonly used dataset for video virtual try-on is the VVT dataset~\cite{dong2019fw}, which contains 791 videos with corresponding garment images.
However, the resolution of the videos is only $256 \times 192$, which limits the networks' ability to learn fine-grained semantic information of clothes.
Moreover, it contains only upper-body clothes, which limits the dataset's applicability to full-body virtual try-on scenarios.
2). The previous works are based on GAN~\cite{goodfellow2014generative}, employing a warping module to obtain the warped clothes aligned with the target pose, which are then fed into a GAN-based generator to generate the final try-on results.
Nevertheless, they heavily rely on the warping module to preserve garment details, since it produces a roughly aligned clothing image to serve as the input for the generator. Once the warping module generates inaccurate warped clothing, the generator will synthesize incorrect results.
The reliance on the warping module for garment detail preservation may introduce limitations in accurately representing clothing characteristics across various body poses and movements.
Furthermore, GAN-based generators inherently exhibit instability during training.

To address the aforementioned issues, we present {\method}: a new dataset consists of 9,700 paired try-on garments and corresponding high-resolution videos ($832 \times 624$), totaling 1,213,694 frames.
This makes {\method} more than 5 $\times$ larger than VVT dataset(totaling 205,675 frames).
What is more, {\method} categorizes garments into three classes: upper-body, lower-body, and dresses.
This categorization holds the promise of significantly improving the performance and applicability of video virtual try-on solutions.

To deal with the drawbacks of GAN-based architectures, we innovatively introduce a novel framework leveraging the powerful diffusion models~\cite{ho2020denoising,rombach2022high,song2021denoising} to tackle the task of video-based virtual try-on.
Diffusion models can offer fine-grained control for generating high-quality content, and the training process is much more stable compared to GANs.
In this paper, we inflate image diffusion models by introducing temporal modules to adapt them to video tasks and we design a Garment Encoder to extract fine-grained semantic information of clothes.
To better integrate the spatial information of garment features with the input video, we propose an attention features fusion mechanism, using the intermediate features of the garment features encoder as conditions for the UNet~\cite{ronneberger2015u}.
Furthermore, we employ a Pose Encoder that takes the dense pose~\cite{guler2018densepose} of the source video as input, aiming to eliminate the influence of background noise while enhancing spatial-temporal consistency.
Through an image-video joint training strategy, our model can synthesize realistic and harmonious videos while preserving the colors, textures, and details of the garments.

Our contributions can be summarized as follows: 1). We propose a novel architecture to address the video virtual try-on task, which leverages the powerful diffusion models to synthesize high-quality try-on videos.
2). We employ Pose Encoder and Temporal Modules to improve temporal consistency. To ensure the generation of high-quality, natural, and realistic try-on videos, we introduce an attention feature fusion mechanism to better exploit the semantic information of garments.
3). We curate a multi-category dataset containing 9,700 pairs of high-quality garment-video samples, making it the largest dataset for the video virtual try-on task to date.
4). Both quantitative and qualitative experiments validate the superiority of our approach.

\section{Related Works}\label{sec:related}
\subsection{Image-based Virtual Try-on}
Given an image of clothing and a photo of a human model, image-based virtual try-on aims to generate an image showcasing the target individual adorned in the designated garment, while preserving the integrity of the areas beyond the clothing, including posture and background. 
Previously, image virtual try-on solution~\cite{han2017viton,wang2018toward,ge2021parser,choi2021viton} was predominantly based on Generative Adversarial Networks (GANs)~\cite{goodfellow2014generative}. 
VITON~\cite{han2017viton} employed a two-stage training strategy, integrating an encoder-decoder generator in the first stage to create a coarse outcome, which was fed into a refinement network, effectively enhancing the garment's adherence to the target body shape while preserving its intricate details.
CP-VTON~\cite{ge2021parser} introduced a Geometric Matching Module to enhance the alignment between the in-shop clothing and the person's representation.
VITON-HD~\cite{choi2021viton} proposed ALIAS to maintain coherence between the clothing and the target try-on area, further improving the resolution of synthesized images to $1024 \times 768$.

Recently, several studies~\cite{Zhu_2023_CVPR_tryondiffusion,chen2023anydoor,morelli2023ladi,baldrati2023multimodal,gou2023taming,kim2023stableviton} have utilized diffusion models to address the challenge of image-based virtual try-on.
LaDI-VITON~\cite{morelli2023ladi} was the first to use latent diffusion models for image-based virtual try-on, applying textual inversion to retain texture information.
Multimodal Garment Designer~\cite{baldrati2023multimodal} introduced multimodal conditions as input for the diffusion model.
StableVITON~\cite{kim2023stableviton} adopted a ControlNet-like~\cite{zhang2023adding} encoder for keeping the fine details of the clothing.
While these approaches have been highly successful in the domain of image try-on, they fall short in accurately capturing the dynamic interaction between clothing and the human body in real-world scenarios.

\subsection{Video-based Virtual Try-on}

The objective of video virtual try-on is to replicate the achievements of image virtual try-on in the realm of videos.
Given a video and an image of target clothing, video virtual try-on aims to transfer the given clothes to a target person in the video.
It demands good temporal consistency between frames in addition to realistic and high-quality try-on outcomes for every frame.
This broadens its potential application scenarios and simultaneously makes it a more challenging research topic.
FW-GAN~\cite{dong2019fw} introduced a flow-navigated module and a warping net to ensure the generated videos have good temporal consistency while maintaining the desired clothing's visual quality.
MV-TON~\cite{zhong2021mv} proposed a two-stage framework that generates a coarse try-on result in the first stage and improves the clothing details through a memory refinement network.
ClothFormer~\cite{jiang2022clothformer} designed a novel warping module to increase the precision of warping operations in garment regions with occlusions.
These works can only manufacture rudimentary apparel and lack clothing intricacies, and they still flicker despite their best attempts.
Furthermore, they only concentrate on upper-body garment try-on.
We argue that the absence of high-quality public datasets is a significant impediment to the development of video try-on. 

\subsection{Video Editing and Image Animation}
Recently, diffusion-based text-driven video editing~\cite{chai2023stablevideo,couairon2024videdit,wu2023tune,qi2023fatezero} has achieved significant success.
It can synthesize videos that align with the provided textual prompts describing an object users wish to edit.
Tune-A-Video~\cite{wu2023tune} fine-tuned a temporal transformer block to correlate the original video with textual clues and used DDIM inversion to preserve areas outside the edited objects. 
FateZero~\cite{qi2023fatezero} proposed an attention maps fusion mechanism to preserve the motion information and structure consistency.
However, it is quite challenging to accurately describe the characteristics of clothing through text alone, especially for garments with complex textures and details.

Given a reference image and a sequence of poses, Image Animation tries to synthesize frames that are consistent with the target poses while maintaining the characteristics of the reference image.
MagicAnimate~\cite{Xu2023MagicAnimateTC} and Animate Anyone~\cite{hu2023animate} both utilized an additional UNet to encode the reference image, successfully synthesizing seamless videos that matched the intended poses.
However, these approaches also altered content outside of the target editing area. For instance, the editing results of Image Animation could not maintain the portions outside of the target clothing unmodified when customers simply wished to try on upper-body clothes.
Moreover, since they take body information (DensePose~\cite{guler2018densepose} or OpenPose~\cite{cao2017realtime}) as conditions, the models are unaware of motions beyond the pose information. For example, when the input video includes facial expressions ({\eg}talking, blinking), Image Animation is unable to reconstruct these facial movements.

\section{{\method} Dataset}\label{sec:dataset}

We argue that a high-quality video virtual try-on dataset should fulfill the following requirements: 
1). It should be publicly available for research purposes. 
2). It should contain paired video-clothing samples, {\ie}an image of clothing and the video of a person wearing the corresponding outfit. 
3). It should feature high resolution to enable models to learn the detailed features of the garments. 
4). It should have enough variety in clothing ({\eg}upper-body, lower-body, and dresses) to accommodate a broad range of application situations.
Unfortunately, there is only one publicly available dataset for video virtual try-on named VVT~\cite{dong2019fw}. 
It contains 791 videos and the corresponding garment images, with a resolution of $256 \times 192$. 
Furthermore, VVT only features garments for the upper body, and the videos are quite repetitive and restricted to catwalks. 
Due to its low resolution and lack of diversity, it has limited the development of video virtual try-on.
To this end, we introduce the {\method} Dataset, a novel dataset for the task of video virtual try-on.
It contains 9,700 clothing-video pairs with a resolution of $832 \times 624$, where the clothing is categorized into three types: upper-body, lower-body, and dresses (including jumpsuits), totaling 1,213,694 frames, which is nearly 6 times the size of the VVT dataset (205,675 frames). 
This makes the {\method} dataset the largest, most diverse, and highest resolution video virtual try-on dataset to date.

{\bf Data collection and annotation.} We first downloaded a variety of clothing images and related video data from the e-commerce website Net-A-Porter~\cite{net-a-porter}.
Each video captures a person wearing an outfit while performing various movements ({\eg}walking, swaying, turning) within an indoor setting, and the images showcase the clothing in a neat, structured form against a pure white background.
However, relying solely on clothing and video data does not meet the needs of try-on tasks.
The primary data for most image try-on tasks can be formulated as follows: 1). An original image of the model, 2). An image of the clothing, 3). A garments-agnostic map, and 4). Information on human body pose.
\begin{wrapfigure}{r}{0.4\linewidth}
     \vspace{-0.2em}
    \includegraphics[width=1\linewidth]{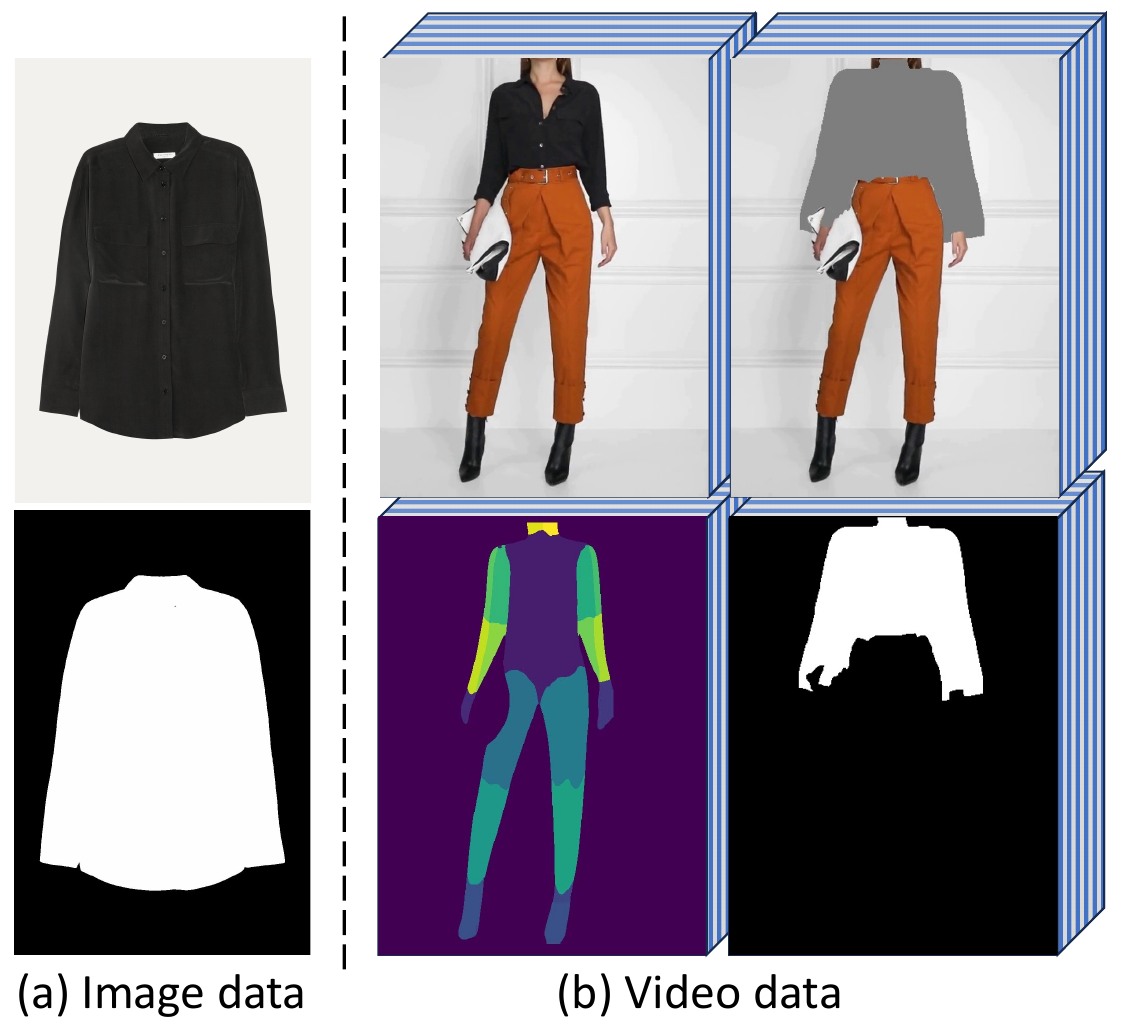}
    \caption{\small{An image-video pair from the {\method} dataset.}}
    \label{fig:datasample}
   \vspace{-0.6em}
\end{wrapfigure}
These signals enable the model to learn how to properly fit the target clothing onto the designated area while providing an awareness of the interaction details between the clothing and the human body.
Therefore, we process the original data in the following steps:
1). We crop the videos to $832 \times 624$ to align with the aspect ratio of the image dataset.
2). To obtain the clothing-agnostic map and the corresponding mask, we use OpenPose~\cite{cao2017realtime} to extract 18 human body keypoints and utilize SCHP~\cite{li2019self} to generate the model's human parsing semantic segmentation mask indicating body parts and the clothing region.
3). We employ DensePose~\cite{guler2018densepose} to extract the pose information, as ShineOn~\cite{kuppa2021shineon} indicates that DensePose is more effective in representation and can better preserve facial details.
4). We utilize SAM~\cite{kirillov2023segany} to calculate masks for the clothing images and use BLIP-2~\cite{li2023blip2} to categorize all clothing into three categories: upper-body, lower-body, and dresses (including jumpsuits).
~\cref{fig:datasample} displays a sample from the {\method} dataset, which includes a video and the associated clothing along with additional information.
\section{Methods}\label{sec:method}

Given a source video \(\boldsymbol{I}_{\text{S}}^{1:N}=[I_S^1, \cdots, I_S^N]\) and a reference garment image \(\boldsymbol{G}_{\text{ref}}\), our goal is to synthesize a realistic, cohesive, and natural try-on video   \(\boldsymbol{I}_{\text{T}}^{1:N}=[I_T^1, \cdots, I_T^N]\) with the person dressed in \(\boldsymbol{G}_{\text{ref}}\) while other areas remain consistent with \(\boldsymbol{I}_{\text{S}}^{1:N}\).
In this section, we initially present the fundamental background of stable diffusion and then elaborate on the details of our method in Sec. \ref{sec:method}, and finally outline our training strategy in Sec. \ref{subsec:training}.

\subsection{Proposed Method}
\label{sec:method}
\noindent\textbf{Stable Diffusion.} Stable Diffusion model~\cite{rombach2022high} is one of the most widely applied latent diffusion models (LDMs) in the community. It consists of a variational autoencoder (VAE)~\cite{kingma2013auto} and a denoising UNet $\epsilon_{\theta}$.
Given an image $x$ and a text prompt $y$, the VAE encoder $\mathcal{E}(\cdot)$ first encodes the image into the latent space, after which a Gaussian noise $\epsilon$ is added to the latent code.
Under the condition of text embedding encoded by the CLIP~\cite{radford2021learning} text encoder, the UNet learns how to predict a Gaussian noise, thereby denoising the noisy latent to a clean one.
Finally, the VAE decoder $\mathcal{D}(\cdot)$ decodes the image representation from the latent space back into the pixel space.
The training loss for the UNet can be formalized as follows:
\begin{equation}
    \mathcal{L}_{LDM} = \mathbb{E}_{\mathcal{E}(\mathbf{x}),\mathbf{y},\epsilon\sim\mathcal{N}(0, 1),t}\left[\lVert\epsilon - \epsilon_{\theta}(\mathbf{z}_t, t, \tau_{\theta}(\mathbf{y}))\rVert_2^2\right],
\end{equation}
where $\mathbf{z}_t$ denotes the noisy latent code at each timestep t, and $\tau_{\theta}(\cdot)$ represents the CLIP text encoder.

\begin{figure}[!h]
  \centering
    \vspace{-1.0cm}
  \includegraphics[width=1.0\linewidth]{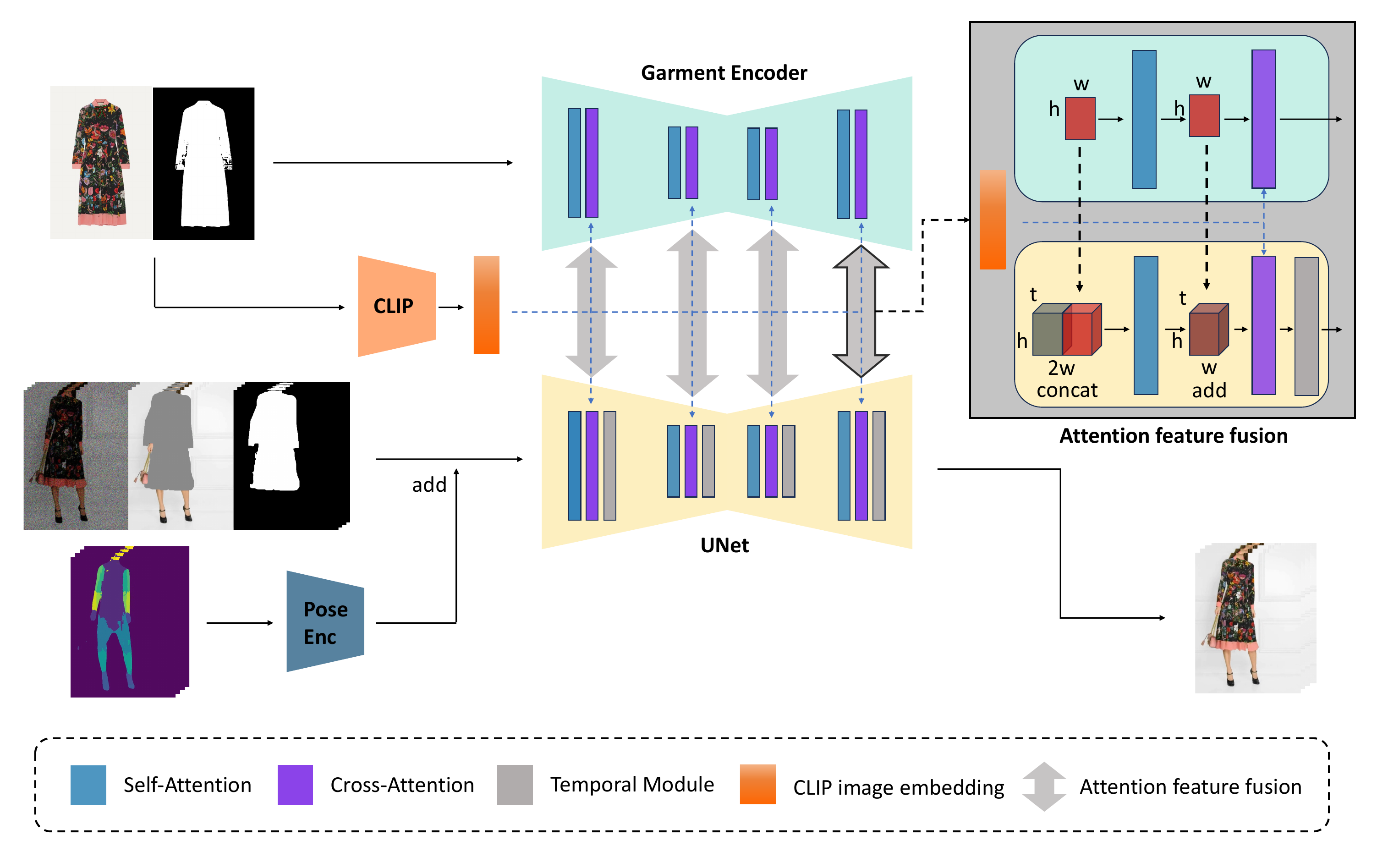}
  \caption{\small{{\bf The overview of {\method}.} First, the noisy video is concatenated with the clothing-agnostic video and the mask video, the pose feature is then added to it. The result serves as input for the UNet. Simultaneously, the Garment Encoder takes the clothing and the mask as input. After that, the attention feature fusion is conducted between the Garment Encoder and the UNet.}}
  \label{fig:pipeline}
  \vspace{-0.5cm}
\end{figure}

{\bf Overview.} The overall framework is presented in \cref{fig:pipeline}. Given a garment image \(\boldsymbol{G}_{\text{ref}}\) $\in\mathbb{R}^{3\times H\times W}$ and the source video \(\boldsymbol{I}_{\text{S}}\) $\in\mathbb{R}^{3\times F\times H\times W}$, accompanied by the clothing-agnostic video \(\boldsymbol{I}_{\text{a}}\) $\in\mathbb{R}^{3\times F\times H\times W}$ and the corresponding mask video \(\boldsymbol{I}_{\text{m}}\) $\in\mathbb{R}^{1\times F\times H\times W}$. We treat the video virtual try-on as a video inpainting problem to adhere the garment onto the clothing-agnostic regions.
Subsequently, we concatenate the noisy video $\mathbf{z}_t$, the latent agnostic video $\mathcal{E}({\boldsymbol{I}_{\text{a}}})$, and the resized mask video $\mathcal{R}({\boldsymbol{I}_{\text{m}}})$ as the input for the UNet.
The original UNet has an input channel size of 4. We expand the initial convolution layer to 9 ({\ie} $4+4+1=9$) channels with weights initialized to zero. 
To model the inter-frame information in the video, we insert temporal modules into the UNet, enabling the model to synthesize temporally consistent results.

Instead of using text to describe the features of the clothing, we employ a Garment Encoder to extract the fine-grained semantic details of the garment.
It copies weights from the UNet and receives a latent clothing image and the resized clothing mask as input.
Simultaneously, we utilize the CLIP image encoder to extract the high-level semantic information of the clothing and inject it into both the UNet and the Garment Encoder.

Furthermore, since the agnostic video eliminates clothing information while also weakening the human pose signals, we additionally design a Pose Encoder $\mathcal{P}(\cdot)$.
It receives the pose sequence \(\boldsymbol{I}_{\text{p}}\) $\in\mathbb{R}^{3\times F\times H\times W}$ extracted by DensePose from the original video and is incorporated into the UNet.

{\bf Garment Encoder.} Most video editing tasks, including video inpainting, use text to describe the regions and contents that are to be edited.
However, text could only provide high-level semantic information, making it difficult to capture the detailed features of the clothing.
And merely replacing the text encoder with the CLIP image encoder doesn't yield satisfactory results either, because CLIP was not trained to model fine-grained details.

To address the aforementioned issues, we design a UNet-like Garment Encoder that captures the fine-grained semantic details of the clothing. 
Through our proposed attention feature fusion mechanism, the model is capable of generating high-fidelity virtual try-on results.
The Garment Encoder inherits the original SD weights, and its initial convolutional layer is initialized with weights of zero for 5 channels.
Next, we explain the attention feature fusion mechanism.
As shown in~\cref{fig:pipeline}, when an intermediate feature map \(\boldsymbol{x}_{\text{g}}\) $\in\mathbb{R}^{h\times w\times c}$ from the Garment Encoder is about to enter the self-attention layer, it is duplicated $t$ times and concatenated with \(\boldsymbol{x}_{\text{d}}\) $\in\mathbb{R}^{t\times h\times w\times c}$ from the UNet along the dimension w, denoted as \(\boldsymbol{x}^{'}_{\text{d}}\) $\in\mathbb{R}^{t\times h\times 2w\times c}$. Then, \(\boldsymbol{x}_{\text{g}}\) and \(\boldsymbol{x}^{'}_{\text{d}}\) undergo self-attention operations, respectively.
We then select the former part of the feature map from the UNet's self-attention layer as the output and, add the outcome from the Garment Encoder's self-attention to it, to further enhance the features of the garment.
Additionally, the CLIP image embedding serves as the condition for the cross-attention layer.

\begin{figure}[!h]
  \centering
  \includegraphics[width=0.9\linewidth]{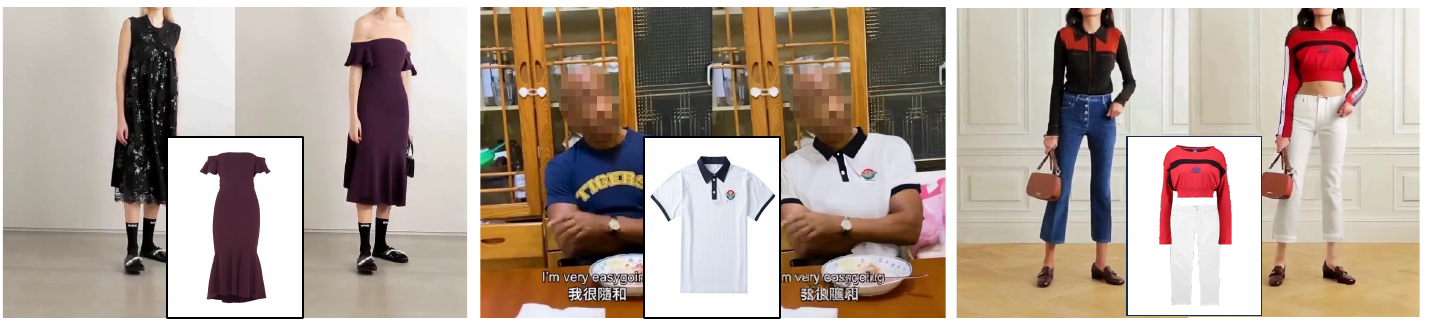}
  \caption{\small{{\method} can handle a variety of clothing.}}
  \label{fig:diversity}
\end{figure}
This design offers the following advantages: 1). The weights of both the Garment Encoder and the UNet are derived from the original SD model, which naturally endows them with SD's capability for image feature extraction. 
2). The attention feature fusion mechanism ensures that the Garment Encoder injects the extracted garment features into the UNet, and the CLIP image encoder provides global semantics of the clothing.

{\bf Pose Encoder.} Since a clothing-agnostic video eliminates the target clothes information in the try-on region, it inevitably loses the human pose signals, which are an essential part for precisely synthesizing a realistic and smooth video.
To this end, we propose a lightweight Pose Encoder to emphasis the pose information from the source video.
It consists of 4 convolutional layers, with a kernel size of 4, strides of 2, and utilizes 16, 32, 64, and 128 channels, respectively.
Specifically, we employ DensePose to extract the pose sequence, which is subsequently input into the Pose Encoder.
The output is then added to the concatenated latent code after passing through the initial convolution layer.
This signal enhances the model's ability to learn how the human body interacts with garments, especially those with complex shapes.

{\bf Temporal Module.} To generate a smooth, spatial-temporal consistent try-on video, we additionally insert Temporal Modules into the original UNet. These modules are designed to capture and integrate the temporal information across frames, leading to a more coherent and fluid video output.
Given a 5D video feature map  \(\boldsymbol{z}\) $\in\mathbb{R}^{b\times c\times f\times h\times w}$, the temporal and spatial axes are reshaped into the batch axis, denoted as  \(\boldsymbol{z}^{'}\) $\in\mathbb{R}^{(b\times h\times w)\times f\times c}$, and then passed through the temporal layers, which conduct self-attention along the dimension f.

\subsection{Training Strategy}
\label{subsec:training}
{\bf Learning objectives.} We employ a single-stage training strategy, conducting joint training on both image and video datasets.
The loss of the entire training process is computed as follows:
\begin{equation}
    \mathcal{L}_{\method} = \mathbb{E}_{\mathcal{E}(\boldsymbol{I}_{\text{cat}}), \mathcal{E}(\boldsymbol{G}_{\text{cat}}), \boldsymbol{I}_{\text{p}}, \boldsymbol{G}_{\text{ref}}, \epsilon\sim\mathcal{N}(0, 1),t}\left[\lVert\epsilon - \epsilon_{\theta}(\mathbf{z}_t, t, \mathcal{P}(\boldsymbol{I}_{\text{p}}), \omega_{\theta}(\mathcal{E}(\boldsymbol{G}_{\text{cat}}), \psi), 	\psi)\rVert_2^2\right],
\end{equation}
\vspace{-1em}

where $\mathcal{E}(\boldsymbol{I}_{\text{cat}})$ represents the concatenation of the noisy video latent code, the clothing-agnostic video latent code, and the resized video mask. $\mathcal{E}(\boldsymbol{G}_{\text{cat}})$ means the concatenation of the clothing latent code and the resized clothing mask. $\omega_{\theta}(\cdot, \cdot)$ denotes the Garment Encoder and $\psi$ represents the CLIP image embedding.

{\bf Image-video joint training.} We train our model on two image datasets, VITON-HD and Dress Code, and one video dataset, {\method}. VITON-HD focuses on upper-body clothing and therefore offers better human pose differentiation and fine details of garment appearance, while Dress Code is larger and can provide a greater variety of clothing.
To combine the advantages of these two image datasets with our {\method} dataset, we design an image-video joint training strategy.
During training, we select a threshold $\lambda$ and draw a random number $r\sim\ U(0, 1)$,  where
$ U(\cdot, \cdot)$ denotes uniform distribution.
When $r \leq \lambda$, we sample N samples from the image datasets, and freeze the temporal module. Otherwise, we extract one video with N frames from the video dataset, and then set the parameters of the temporal module to be trainable.
With this training strategy, our model can better learn garment details while also effectively modeling the temporal dimension.

\begin{figure}[!h]
  \centering
  \includegraphics[width=0.8\linewidth]{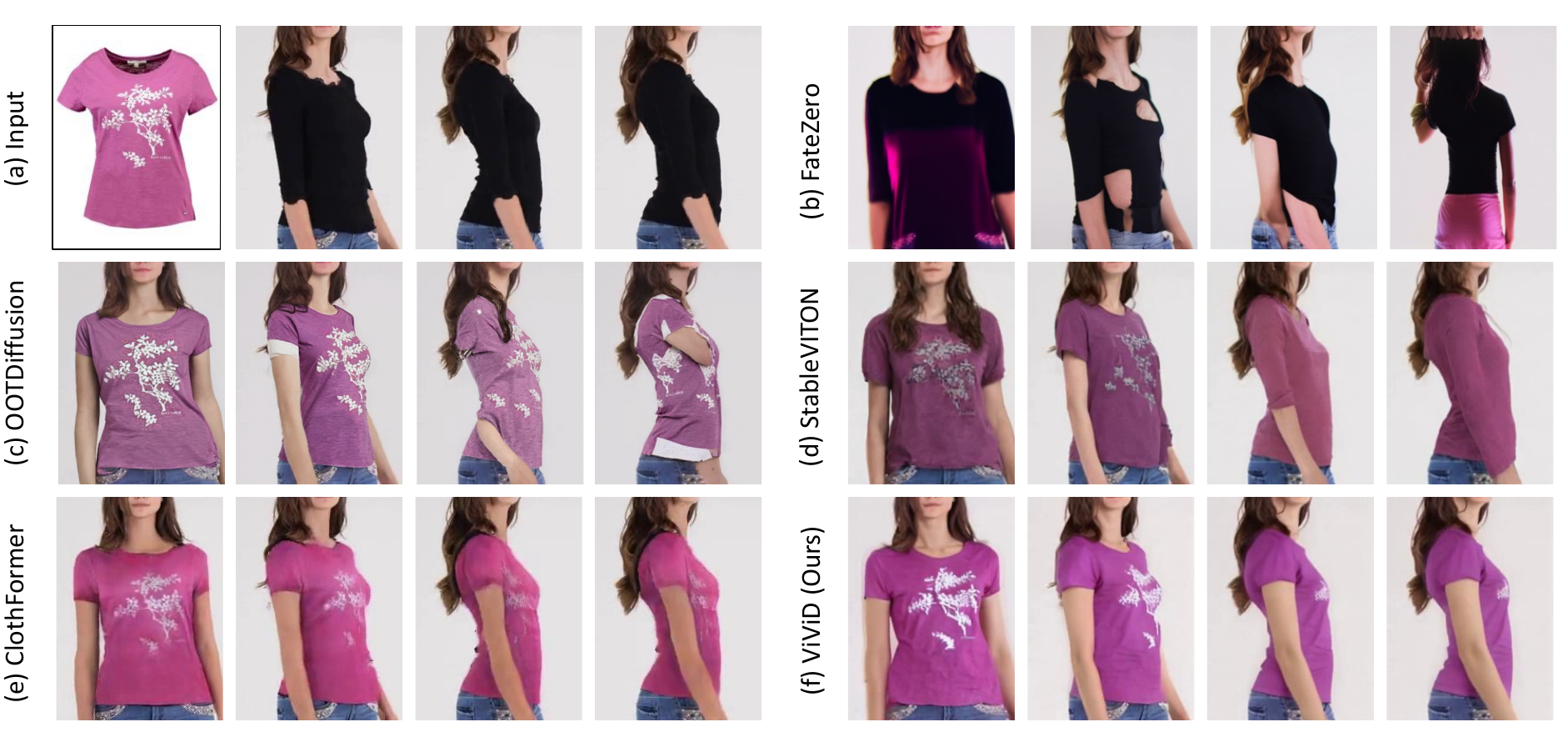}
  \caption{\small{Qualitative comparison results of our {\method} with other visual try-on solutions on the VVT dataset.}}
  \label{fig:comparison}
  \vspace{-0.5cm}
\end{figure}

\section{Experiments}\label{sec:exp}

\subsection{Datasets} 
We train our model on two image datasets, VITON-HD~\cite{choi2021viton} and Dress Code~\cite{morelli2022dresscode}, and one video dataset, {\method}. 
We do not use the VVT~\cite{dong2019fw} dataset due to its low resolution, limited variety of clothing types, and simple motions, which make it not conducive for the model to learn clothing detail representations and temporal relationships in videos.
The {\method} dataset contains 9,700 video-image pairs, with a resolution of $832 \times 624$.
We divide it into 7,759 videos for the training set and 1,941 videos for the test set, totaling 1,213,694 frames, which is almost 6 times larger than the VVT dataset.
The dataset will be available for research purposes.

\subsection{Implementation Details}
We initialize the UNet and Garment Encoder using the weights from Stable Diffusion-1.5, the temporal module is initialized with the weights from the motion module of AnimeDiff~\cite{guo2023animatediff}, and the CLIP image encoder is borrowed from Image Variations~\cite{imagevariations}.
During training, all data is resized to a uniform resolution of $512 \times 384$.
We train our model on 4 Nvidia A100 GPUs, using a learning rate of 1e-5, for approximately 120 hours.
We employ an image-video joint training strategy.
Upon selection of the image datasets, we randomly select 24 images and subsequently freeze the temporal module. 
Conversely, when the video dataset is chosen, we sample a continuous sequence of 24 frames, during which the temporal module is activated.

\subsection{Qualitative Results}
\begin{wrapfigure}{r}{0.38\linewidth}
    \vspace{-2.5em}
    \includegraphics[width=1\linewidth]{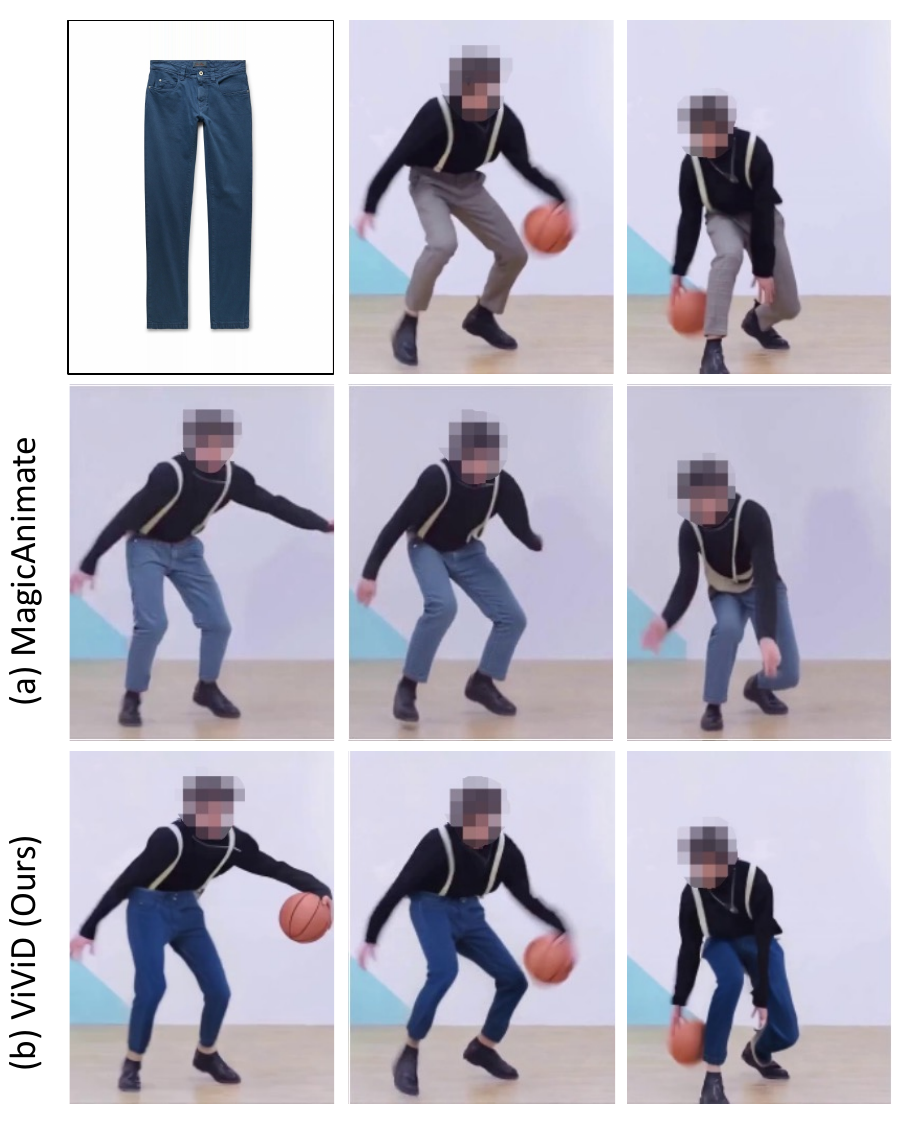}
    \caption{\small{The man lost his ball\includegraphics[width=0.07\linewidth]{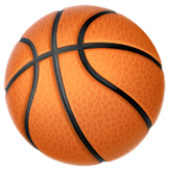}} in MagicAnimate\includegraphics[width=0.07\linewidth]{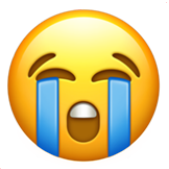}.}
     \vspace{-1.0em}
    \label{fig:image_animation}
\end{wrapfigure}
We conduct a comparison with existing visual try-on methods on the VVT dataset, including a video editing solution (FateZero~\cite{qi2023fatezero}), two image-based solutions (StableVITON~\cite{kim2023stableviton}, OOTDiffusion~\cite{xu2024ootdiffusion}), and a video-based try-on solution (ClothFormer~\cite{jiang2022clothformer}).
Note that the VVT dataset is not part of our training data.
As shown in~\cref{fig:comparison}, (a) represents the given input, including a T-shirt and a video clip. (b) illustrates that FateZero fails to transfer the T-shirt onto the person and results in significant artifacts.
(c) OOTDiffusion and (d) StableVITON successfully synthesize decent try-on results from a frontal pose perspective, yet they struggle with temporal consistency, particularly when the model turns.
In comparison to the aforementioned methods, (e) ClothFormer is capable of synthesizing try-on videos with better temporal consistency. 
However, when the woman turns around, the floral pattern on the clothing undergoes implausible deformation and flickering.
Our method (f) {\method} can generate high-quality outcomes that preserve the details of the clothing while maintaining temporal consistency without producing artifacts, even during actions with large movements such as turning.
We also compare our method with Image Animation approach (MagicAnimate~\cite{Xu2023MagicAnimateTC}). We select one frame from the video and then use OOTDiffusion to generate the try-on image, which we propagate throughout the entire video. 
As shown in~\cref{fig:image_animation}, MagicAnimate fails to preserve the content outside of the target try-on area.
Additionally, our model is capable of handling a variety of different scenarios. 
Previous video virtual try-on models could only deal with tops. 
As illustrated in ~\cref{fig:diversity}, our model can adapt to a variety of clothing types.

\subsection{Quantitative Results}
    
    
\begin{table}
\setlength\tabcolsep{2pt}%
  \centering
  \vspace{-2mm}
    \caption{Comparison on the VVT dataset: $\uparrow$ denotes higher is better, while $\downarrow$ indicates lower is better.}
    \vspace{2mm}
    
    \resizebox{0.8\linewidth}{!}{
   \begin{tabular}{lccccc}
    \toprule
    Method  & SSIM$\uparrow$ & LPIPS$\downarrow$ & $VFID_{I3D}\downarrow$ & $VFID_{ResNeXt}\downarrow$ \\
    \midrule
    CP-VTON~\cite{wang2018toward} &0.459 &0.535 &6.361 &12.10 \\
    FW-GAN~\cite{dong2019fw}  &0.675 &0.283 &8.019  &12.15  \\
    PBAFN~\cite{ge2021parser}  &0.870 &0.157 &4.516  &8.690  \\
    ClothFormer~\cite{jiang2022clothformer} &{0.921} &0.081 &{3.967} &{5.048}\\
    
    \textbf{\method}  & \textbf{0.949} & \textbf{0.068} & \textbf{3.405}&\textbf{5.074} \\
    \bottomrule
  \end{tabular}
}
  \label{tab1: vvt}
\end{table}
We conduct the quantitative comparisons between {\method} and other approaches on the VVT dataset. 
We use the structural similarity (SSIM)~\cite{wang2004image} and the perceptual image patch similarity (LPIPS)~\cite{zhang2018unreasonable} to evaluate the image quality. 
For video results, we leverage two backbones: I3D~\cite{carreira2017quo} and 3D-ResNetXt101~\cite{hara2018can} to compute the Video Frechet Inception Distance (VFID) to measure the visual quality and temporal consistency.
As demonstrated in Table~\ref{tab1: vvt}, {\method} surpasses previous image-based methods and video-based methods. 

\subsection{Ablation Study}

To verify the effectiveness of our Garment Encoder and image-video joint training strategy, we conduct two ablative experiments.

\begin{figure}[!h]
  \centering
  \includegraphics[width=0.5\linewidth]{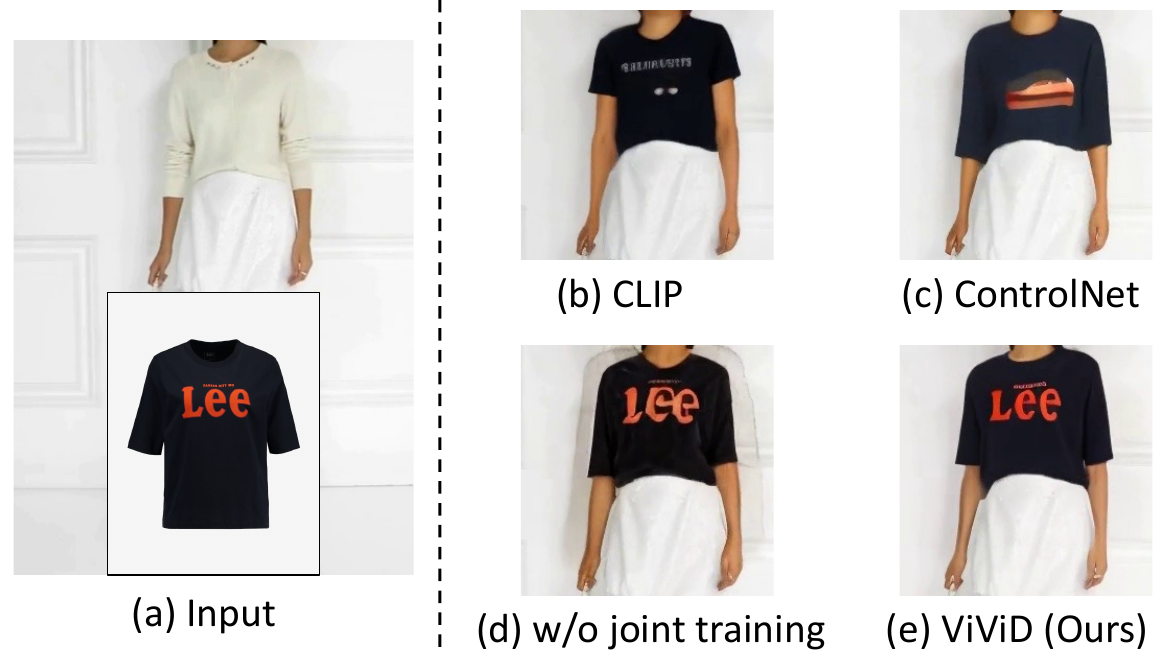}
  \caption{\small{Ablations for the Garment Encoder and the joint training strategy.}}
  \label{fig:ablation}
  \vspace{-0.5cm}
\end{figure}

{\bf Garment Encoder.} To illustrate the Garment Encoder's efficiency in extracting clothing representations, we conduct the following two experiments: 1) replace it with ControlNet~\cite{zhang2023adding}, 2) remove it completely.
As illustrated in~\cref{fig:ablation} (b) and (c), both replacing the Garment Encoder with ControlNet and completely removing it fail to yield outcomes with fine details.
This indicates that our Garment Encoder and attention feature fusion mechanism are capable of preserving fine-grained semantic information of clothing.

{\bf Image-video joint training.} We exploit the benefits of both image and video datasets through our image-video joint training strategy.
As shown in~\cref{fig:ablation} (d), instead of using 
%
a joint training strategy, employing a two-stage training strategy (where the first stage is trained on image datasets and the second stage is trained on the video dataset) results in the loss of clothing details. 
This is because, during the transition from the first to the second stage, the shift in data format and context prevents the model from effectively retaining and applying the intricate texture and style features of the clothing learned during the image-focused training, leading to a degradation in detail preservation in video output.

\section{Conclusions}\label{sec:conclusion}
In this paper, we present {\method}, an innovative framework that utilizes powerful diffusion models for solving video virtual try-on challenges. 
Comprehensive experiments demonstrate that {\method} can produce video try-on results with high visual quality and temporal consistency. 
We also collect a new dataset, which is the largest dataset for this task, featuring garments of multiple categories and high-resolution image-video pairs. 
We believe that our methodology and dataset can serve as valuable references for researchers in the field of video virtual try-on.

{\small
\bibliographystyle{plainnat}
\bibliography{ref}
}

\newpage
\appendix
\section*{Appendix}\label{sec:appendix}

\section{More implementation details} 

\textbf{Training and Inference details.} AdamW~\cite{loshchilov2017AdamW} is chosen as the default optimizer, with a learning rate set to $1 \times 10^{-5}$.
During the training process, all the data undergoes resizing to a resolution of $512 \times 384$ pixels. When the video dataset is selected, we sample a 24-frame video clip with the sample rate set to 4. We train our model in 4 Nvidia A100 GPUs (80GB), the total training steps is 60,000, with fp16 mixed precision.

We train the Temporal Module using 24-frame video clips due to the VRAM capacity, which leads to difficulties in generating longer try-on results. To tackle this challenge, we employ a sliding window method, following MagicAnimate~\cite{Xu2023MagicAnimateTC}, to ensure long-range video consistency. Specifically, we divide a long input video into several segments with overlap, then we denoise every single segment, conduct noise interpolation in the overlapped parts, and finally combine each segment to generate a spatial-temporal consistent long try-on video.

\textbf{Data Augmentation.} We apply data augmentations for the image and video datasets. We use horizontal flip (with probability 0.5), random
affine shifting and scaling (limit of 0.2, with probability 0.5) to the clothing data ({\ie}clothing image and the corresponding mask) and video data ({\ie}source video, pose video, clothing-agnostic video and the corresponding mask video). Note that we only apply spatial augmentations, as we argue that different illumination conditions in videos can cause the color deviation of the garments.

\section{More experimental results} 
\begin{figure}[!h]
  \centering
  \includegraphics[width=0.8\linewidth]{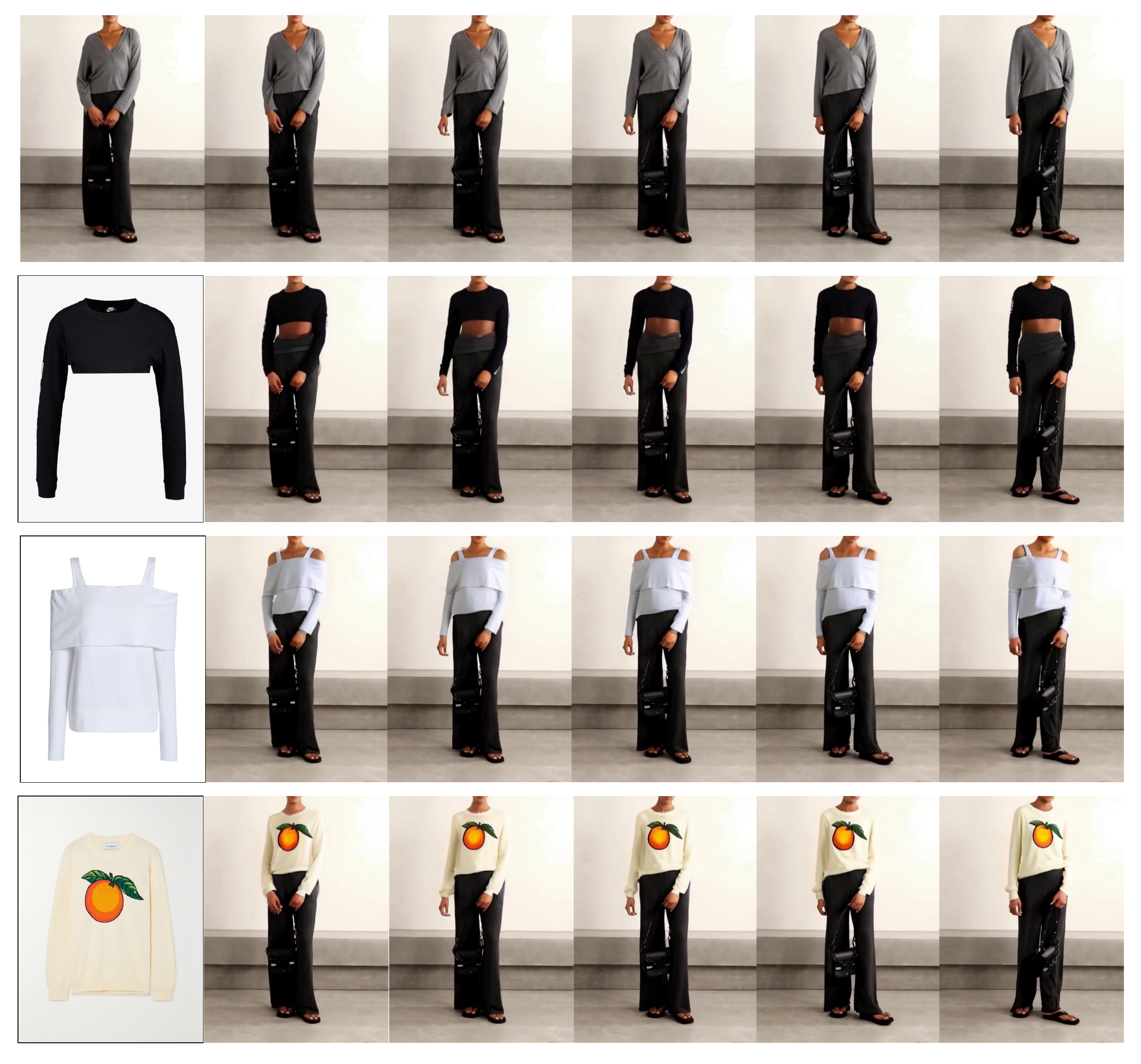}
  \caption{\small{More try-on results generated by {\method}}.}
  \label{fig:more1}
  \vspace{-0.5cm}
\end{figure}

\begin{figure}[!h]
  \centering
  \includegraphics[width=0.96\linewidth]{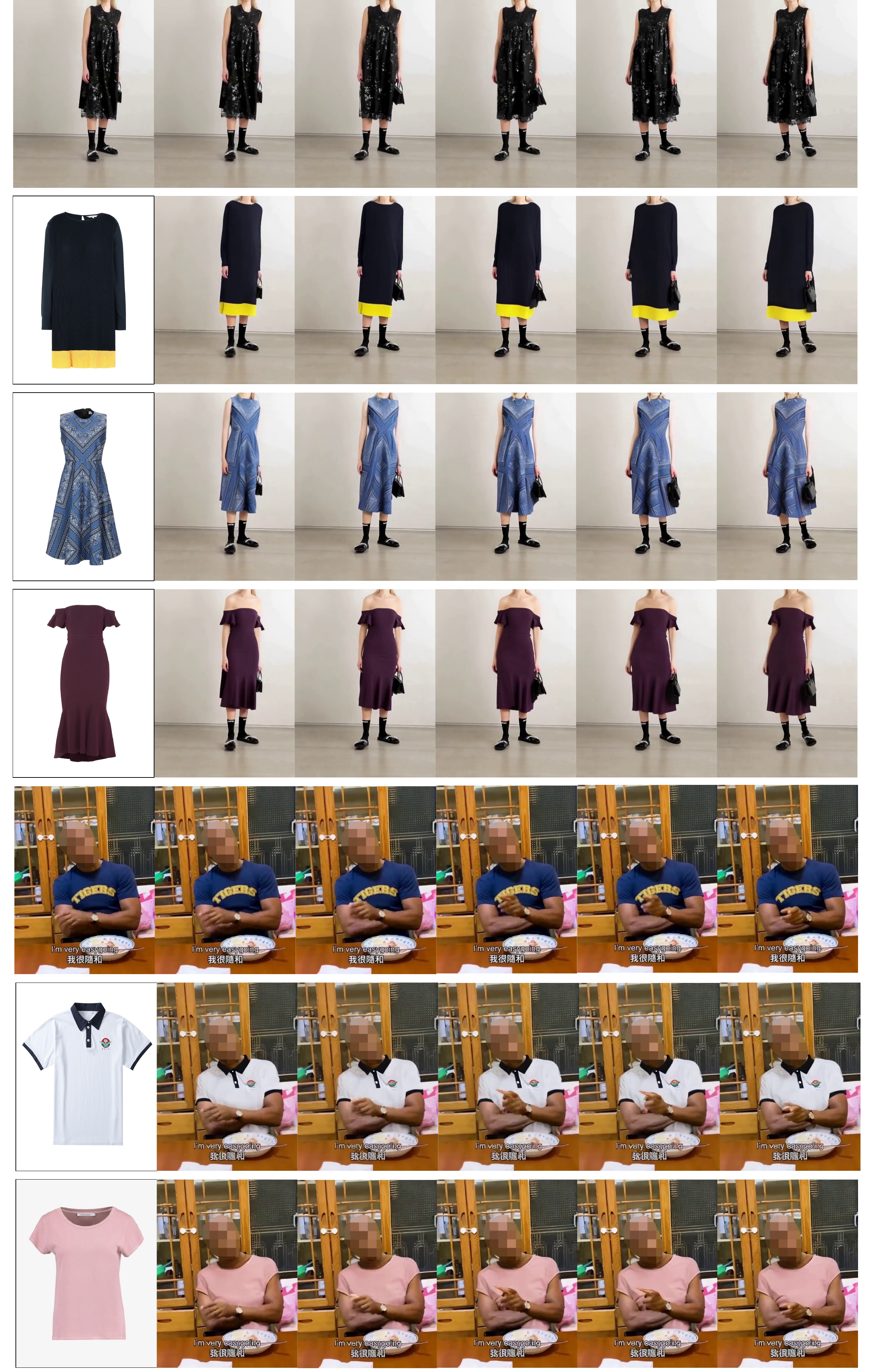}
  \caption{\small{More try-on results generated by {\method}. The 1st and 5th rows are source videos.}}
  \label{fig:more2}
\end{figure}

\begin{figure}[!h]
  \centering
  \includegraphics[width=0.96\linewidth]{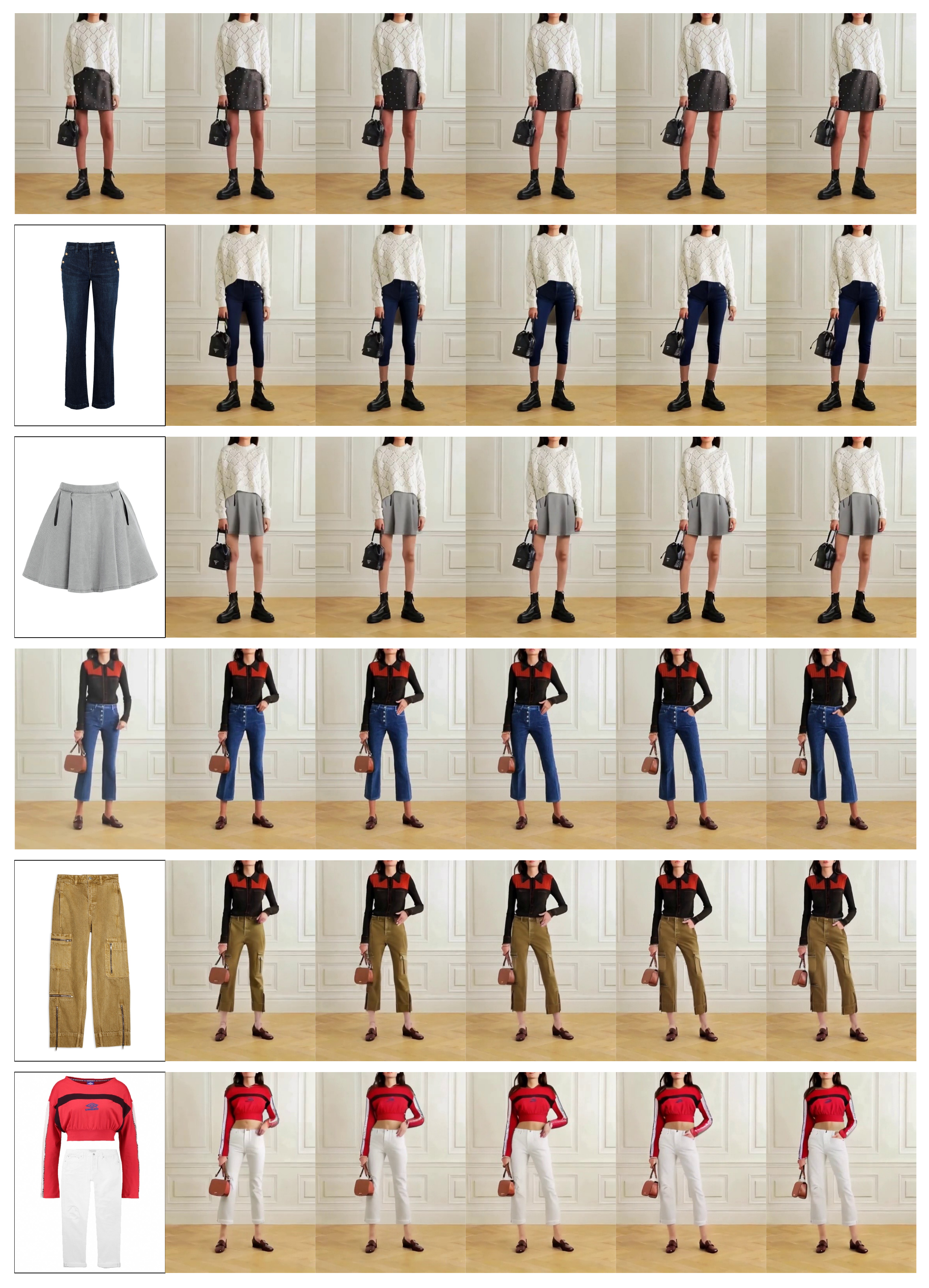}
  \caption{\small{More try-on results generated by {\method}. The 1st and 4th rows are source videos.}}
  \label{fig:more3}
  \vspace{1.5cm}
\end{figure}

As shown in \cref{fig:more1}, \cref{fig:more2} and \cref{fig:more3}, our approach can be applied to a wide variety of clothing scenarios.


\section{Discussion} 

\begin{figure}[!h]
  \includegraphics[width=0.95\linewidth]{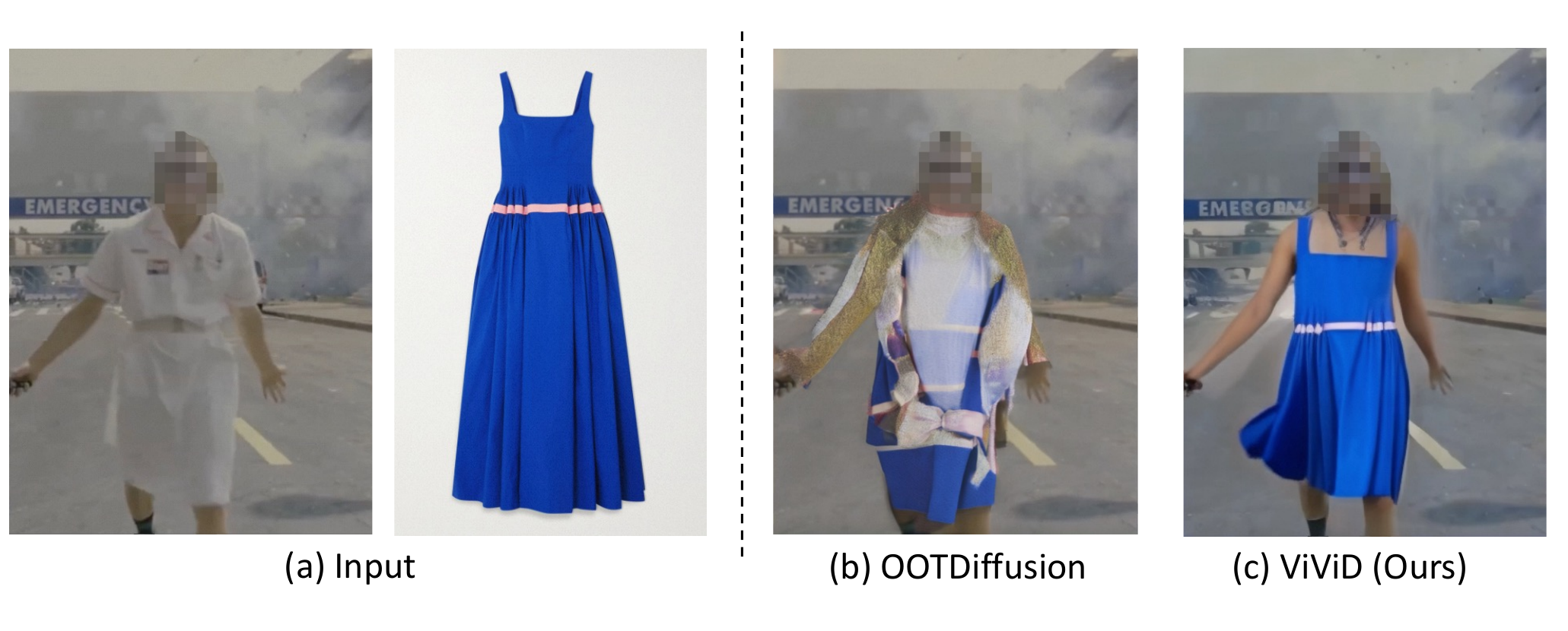}
  \caption{\small{{\method} exhibits artifacts when dealing with outdoor scenarios.}}
  \label{fig:limitation}
\end{figure}

\textbf{Limitations.} Due to the ViViD dataset's focus on indoor scenes, our model exhibits noticeable artifacts when dealing with outdoor scenarios, as illustrated in \cref{fig:limitation}. However, similar issues also exist in other try-on methods like OOTDiffusion\cite{xu2024ootdiffusion}. We believe the community still requires a video try-on dataset featuring in-the-wild videos to further broaden the application spectrum of video virtual try-on.

\textbf{Potential societal impact.} Our model is capable of providing a more authentic try-on experience, allowing consumers to more accurately assess the style and size of clothing. This has the potential to reduce the rate of returns within e-commerce, which is favorable for sustainable development and environmental protection. Additionally, designers can employ this technology to visualize their innovation, while consumers may find the process of customizing personalized apparel more accessible.

However, to achieve accurate virtual try-on effects, it may be necessary to collect user body measurements and shape data, potentially leading to privacy breaches. Furthermore, video virtual try-on technology could be misused to create deceptive video content, raising ethical and legal concerns. Therefore, it is necessary to refine the relevant legal regulations and to guide and regulate the use of the models.

\end{document}